\documentclass{bmvc2k}

\usepackage[acronym]{glossaries}
\usepackage[percent]{overpic}
\usepackage{boldline}
\usepackage{booktabs}
\usepackage{hyperref}

\title{MS-ASL: A Large-Scale Data Set and Benchmark for Understanding \\ American Sign Language}

\addauthor{Hamid Reza Vaezi Joze}{hava@microsoft.com}{1}
\addauthor{Oscar Koller}{Oscar.Koller@microsoft.com}{2}

\addinstitution{
 Microsoft\\
 Redmond, WA, USA
}
\addinstitution{
 Microsoft\\
 Munich, Germany
}

\runninghead{Vaezi Joze, Koller}{MS-ASL}

\def\etal{\emph{et al}\bmvaOneDot}

\begin{document}

\maketitle

\begin{abstract}

Sign language recognition is a challenging and often underestimated
problem comprising multi-modal articulators (handshape, orientation,
movement, upper body and face) that integrate asynchronously on
multiple streams.
Learning powerful statistical models in such a scenario requires much
data, particularly to apply recent advances of the field. 
However, labeled data is a scarce resource for sign language due to
the enormous cost of transcribing these unwritten languages.

We propose the first real-life large-scale sign language data set
comprising over 25,000 annotated videos, which we thoroughly evaluate
with state-of-the-art methods from sign and related action recognition.
Unlike the current state-of-the-art, the data set allows to
investigate the
generalization to unseen individuals (signer-independent test) in a 
realistic setting with over 200 signers. 
Previous work mostly deals with limited vocabulary tasks, while here,
we cover a large class count of 1000
signs in challenging and
unconstrained real-life recording conditions.
We further propose I3D, known from video classifications, as a
powerful and suitable architecture for sign language recognition,
outperforming the current state-of-the-art by a large margin. The data set is publicly available to the community.
\end{abstract}

\vspace{-10pt}
\section{Introduction}
In the US, approximately 
around 500,000 people use American sign language (ASL) as primary
means of communication~\cite{mitchell2006many}. ASL is also used in
Canada, Mexico and 20 other countries. Just like any other natural language,
it covers its unique vocabulary as well as a grammar which is
different from spoken English.
We are intrigued by sign language and accessibility for the Deaf and believe
sign recognition is an exciting field that offers many
challenges for computer vision research.  

For decades, researcher from different fields have tried to solve the challenging problem of sign language recognition. Most of the proposed approaches
rely on external devices such as additional
RGB~\cite{brashear2003using} or depth cameras~\cite{uebersax2011real,
  zafrulla2011american}, sensor~\cite{liang1998real,mehdi2002sign} or
colored gloves~\cite{wang2009real}. However, such requirements limit
the applicability to specific settings where such resources are
available. Opposed to that, non-intrusive and purely vision based sign
recognition will allow for general usage. 
With the appearance of deep learning based methods and their powerful
performance on computer vision tasks, the requirements on training
data have changed dramatically from few hundred to thousands of samples being needed to train strong
models. Unfortunately, public large scale sign language resources
suitable for machine learning are very limited and there is currently
no public ASL data set big enough to evaluate recent deep
learning approaches.
This prevents recent computer
vision trends to be applied to this field. 
As such, our goal is to advance the sign language recognition
community and the related state-of-the-art by releasing a new large-scale data
set, establishing thorough baselines and carrying over recent computer
vision trends. With this work, we make the following contributions: 
    (1) We release the first large-scale ASL data set called
      MS-ASL that covers over 200 signers, signer independent sets,
      challenging and unconstrained recording conditions and a large
      class count of 1000 signs ~\footnote{Instructions and download links: \href{https://www.microsoft.com/en-us/research/project/ms-asl/}{https://www.microsoft.com/en-us/research/project/ms-asl/}}.
    (2) We evaluate current state-of-the-art approaches: 2D-CNN-LSTM,
    body key-point, CNN-LSTM-HMM and 3D-CNN as baselines.
    (3) We propose I3D (known from action recognition) as a powerful and
  suitable architecture for sign language recognition that outperforms
  previous state-of-the-art by a large margin and provide new
  pre-trained model for it. 
    (4)  We estimate the effect of number of classes and number of samples on the performance.



\section{Previous Works}
\label{sec:relatedwork}
\textbf{Recognition methods:} Researchers have tried to solve the challenges of sign language recognition in different ways. 
In 1983, the first work was a glove based device that allowed to recognize ASL
fingerspelling based on a hardwired circuit~\cite{grimes_digital_1983}.
In the meantime, there have been a lot of related approaches which
rely on tracked hand movements based on sensor gloves for sign
recognition~\cite{charayaphan_image_1992,fels_glovetalk_1993,liang1998real,mehdi2002sign,oz2011american}.  
Some works extended this by adding a camera as a new source of
information~\cite{mitchell2006many} and they showed that adding video
information improves the accuracy of detection but the method mainly
relies on the glove sensors. 

In 1988, Tamura~\etal 
\cite{tamura_recognition_1988} were the first to follow vision-based sign language recognition. They built a system to recognize 10
isolated signs of Japanese sign language using simple color
thresholding. 
Because the sign is performed in 3-dimensions, many vision based
approaches use depth
information~\cite{kuznetsova2013real,uebersax2011real,
  zafrulla2011american} or multiple
cameras~\cite{brashear2003using}. Some rely on colored gloves to ease
hand and finger tracking~\cite{wang2009real,cooper_sign_2010}. 
In this paper we focus on non-intrusive sign language recognition
using only a single RGB camera as we believe this will allow to design
tools for general usage to empower everybody to communicate with a
deaf person using ASL. 
The sole use of RGB for sign detection is not new, traditional
computer vision techniques particularly with Hidden Markov
Models~\cite{starner1995visual,starner1997real,vonagris_recent_2008,forster_modality_2013},
mainly inspired by improvements in speech recognition, have been in
use in the past two decades. With the advances of deep learning and
convolutional networks for image processing the field has evolved
tremendously. Koller~\etal showed large improvements embedding 2D-CNNs
in HMMs~\cite{koller16:hybridsign,koller_deep_2018}, related works
with 3D-CNNs exist~\cite{huang2015sign,camgoz2016:3dconv} and weakly supervised
multi-stream modeling is presented in~\cite{koller_weakly_2019}. 
However, sign language recognition still lags behind related fields in
the adoption of trending deep learning architectures. To the best of
our knowledge no prior work exists that leverages latest findings from
action recognition with I3D networks or complete body key-points which
we will address with this paper. 

\textbf{Sign language data sets:}
Some outdated reviews of sign language corpora exists~\cite{loeding_progress_2004}. Below,
we have reviewed sign language data sets with explicit setups intended
for reproducible pattern recognition research. 
The \textbf{Purdue RVL-SLLL ASL database}
\cite{kak_purdue_2002,wilbur_purdue_2006} contains 10 short stories with a vocabulary of 104 signs and a total
sign count of 1834  produced by 14 native
signers in a  lab environment under controlled lighting. 
%
The RWTH-BOSTON corpora were originally created for linguistic research~\cite{athitsos2008american} and packaged for pattern recognition purposes later. 
The \textbf{RWTH-BOSTON-50}~\cite{zahedi05:dagm05} and the \textbf{RWTH-BOSTON-104}
corpus contain isolated sign language with a vocabulary of 50
and 104 signs. The \textbf{RWTH-BOSTON-400} corpus contains a
vocabulary of 483 signs and also constitutes of continuous signing by 5
signers. 
The \textbf{SIGNUM
corpus}~\cite{von_agris_significance_2008} provides two evaluation sets: first a multisigner set with 25 signers,
each producing 603 predefined sentences with 3703 running gloss
annotation and a vocabulary of 455 different signs.  Second, it has a
single signer setup where the signer produces three repetitions of the
given sentences.
%
In the scope of the DictaSign project, multi-lingual sign language resources have
been created~\cite{braffort_sign_2010,efthimiou_sign_2012,matthes_dictasignbuilding_2012}. 
However, the produced corpora are not well curated and made available
for reproducible research. 
The \textbf{Greek Sign
Language (GSL) Lemmas Corpus}~\cite{efthimiou_gslc_2007} constitutes 
such a data collection. It provides a subset with isolated sign
language (single signs) that contains 5 repetitions of the
signs produced by two native signers. However, different versions of
this have been used in the literature disallowing fair
comparisons and the use as benchmark corpus. The corpus has been
referred to with
1046
signs~\cite{pitsikalis_advances_2011,theodorakis_dynamicstatic_2014},
with 984 signs~\cite{cooper_reading_2011} and  with
981 signs~\cite{ong_sign_2014}.
Additionally, a continuous 100 sign version of the data set has been
used in~\cite{roussos_dynamic_2013}. The reason for all these
circulating subsets is that the data has not
been made publicly available.
\textbf{DEVISIGN} is a Chinese sign language data set featuring isolated single
signs performed by 8 non-natives~\cite{chai_devisign_2014} in a laboratory environment (controlled background). The data
set is organized in 3 subsets,
covers a vocabulary of up to 2000 isolated signs and provides RGB with
depth information in 24000 recordings.
A more recent data set~\cite{huang_videobased_2018} covers 100 continuous chinese sign language sentences produced five times by 50 signers. It has a vocabulary of 178 signs. It can be considered as staged recording.
The \textbf{Finish  S-pot}  sign spotting
task~\cite{viitaniemi_spota_2014} is based on the controlled recordings from the Finish sign language
lexicon~\cite{finishassociationofthedeaf_suvi_2015}. It covers 1211 isolated citation form signs that need to be spotted in
 4328 continuous sign language videos. However, the task has not been widely adopted by the field.
The \textbf{RWTH-PHOENIX-Weather 2014}~\cite{forster_extensions_2014,koller_continuous_2015}
and \textbf{RWTH-PHOENIX-Weather 2014 T}~\cite{camgoz_neural_2018} are
large scale real-life sign language corpora that feature professional
interpreters recorded from broadcast news. They cover continuous
German sign language with a vocabulary of over 1000 signs, about 9
hours of training data. It only features 9 signers and limited computer vision
challenges. 
There are several groups which experimented with their own private data
collection resulting in corpora with quite limited size in terms of
total number of annotations and vocabulary such as \textbf{UWB-07-SLR-P corpus of Czech sign language}~\cite{campr_collection_2008}, data set by Barabara Loedings
Group~\cite{loeding_progress_2004,nayak_finding_2012} and small scale corpora \cite{bosphorusSign,fagiani_new_2012}.


\begin{table}[h]
\small
\begin{center}
  \renewcommand{\arraystretch}{0.9}
  \setlength{\tabcolsep}{1.5pt}
\begin{tabular}{ l rrcrrc}
\toprule
Data set        & classes & signer & independ. & videos & real-life \\
\midrule
  Purdue ASL~\cite{wilbur_purdue_2006}    & 104     & 14     & no        & 1834   & no        \\
Video-Based CSL~\cite{huang_videobased_2018} & 178     & 50     & yes        & 25000  & no        \\  
Signum~\cite{von_agris_significance_2008}          & 465     & 25     & yes       & 15075  & no        \\
RWTH-Boston~\cite{zahedi05:dagm05}     & 483     & 5      & no        & 2207   & no        \\  
  RWTH-Phoenix~\cite{forster_extensions_2014}  & 1080    & 9      & no        & 6841   & yes       \\
Devisign~\cite{chai_devisign_2014}        & 2000    & 8      & no        & 24000  & no        \\
  \midrule
This work       & 1000    & 222    & yes       & 25513  & yes       \\
  \bottomrule
\end{tabular}
\caption{Comparison of public sign language data sets.} 
\label{tab:compare-datasets}
\end{center}
\vspace{-13pt}
\end{table}
Table~\ref{tab:compare-datasets} summarizes the mentioned data sets.
To the best of our knowledge RWTH-PHOENIX-Weather 2014 and DEVISIGN
are currently the only publicly available data sets that are large
enough to cover recent deep learning approaches. However, both data
sets are lacking the variety and number of signers to advance the
state-of-the-art with respect to the important issue of signer
independence and computer vision challenges from natural unconstrained
recordings. 
In the scope of this work, we propose the first ASL data set
that covers over 200 signers, signer independent sets, challenging and
unconstrained recording conditions and a large class count of 1000 gloss level
signs. 

\vspace{-10pt}
\section{Proposed ASL Data Set}
\label{sec:proposedDS}
Since there is no public ASL data set suitable for large-scale sign language
recognition, we looked for realistic data sources. The
deaf community actively uses public video sharing platforms for
communication and study of ASL. Many of those videos are
captured and uploaded by ASL students and teachers. They constitute
challenging material with large variation in view, background,
lighting and positioning. Also from a language point of view, we
encounter regional, dialectal and inter-signer variation. This seems
very appealing from a machine learning point of view as it may further
close the gap in learning signer independent recognition systems that
can perform well in realistic circumstances. Besides having access to
well suited data, the main issue remains labeling which requires skilled ASL natives. 

We noticed that a lot of the public videos have manual subtitles,
captions, descriptions or a video title that indicates which signs are
being performed in it.  
We therefore decided to access the public ASL videos and obtain the text from all those sources. 
We process these video clips automatically in three distinct ways:
(1) For longer videos, we used Optical Character Recognition (OCR) to find printed labels and their time of occurrence. 
(2) Longer videos may contain video captions that provide the sign descriptor and the temporal segmentation.
(3) In short videos we obtained the label directly from the title.

In the next step, we detected bounding boxes and used face recognition
to find and track the signer. This allowed identification of
descriptions that refer to a static image rather than an actual
signer. If we identified multiple signers performing one after the
other, we  splited the video up into smaller samples. 

In total we accessed more than 45,000 video samples that include words
or phrases in their descriptions. We sorted the words based on
frequency to find the most frequently used ones while removing
misspellings and OCR mistakes. Since many of the ASL vocabulary
publicly accessible videos belong to teachers performing a lesson
vocabulary or students doing their homework, all top hundred words
belong to ASL tutorial books~\cite{zinza2006master,
  vicars2012american} vocabulary units. 
  Some of the videos referenced in MS-ASL originate from ASL-LEX~\cite{caselli_asllex_2017}.

\subsection{Manual Touch-up}
\label{sec:manual-touch-up}

\begin{figure}[t]
\begin{minipage}{0.45\textwidth}
   \centering
   \includegraphics[width=1.0\linewidth]{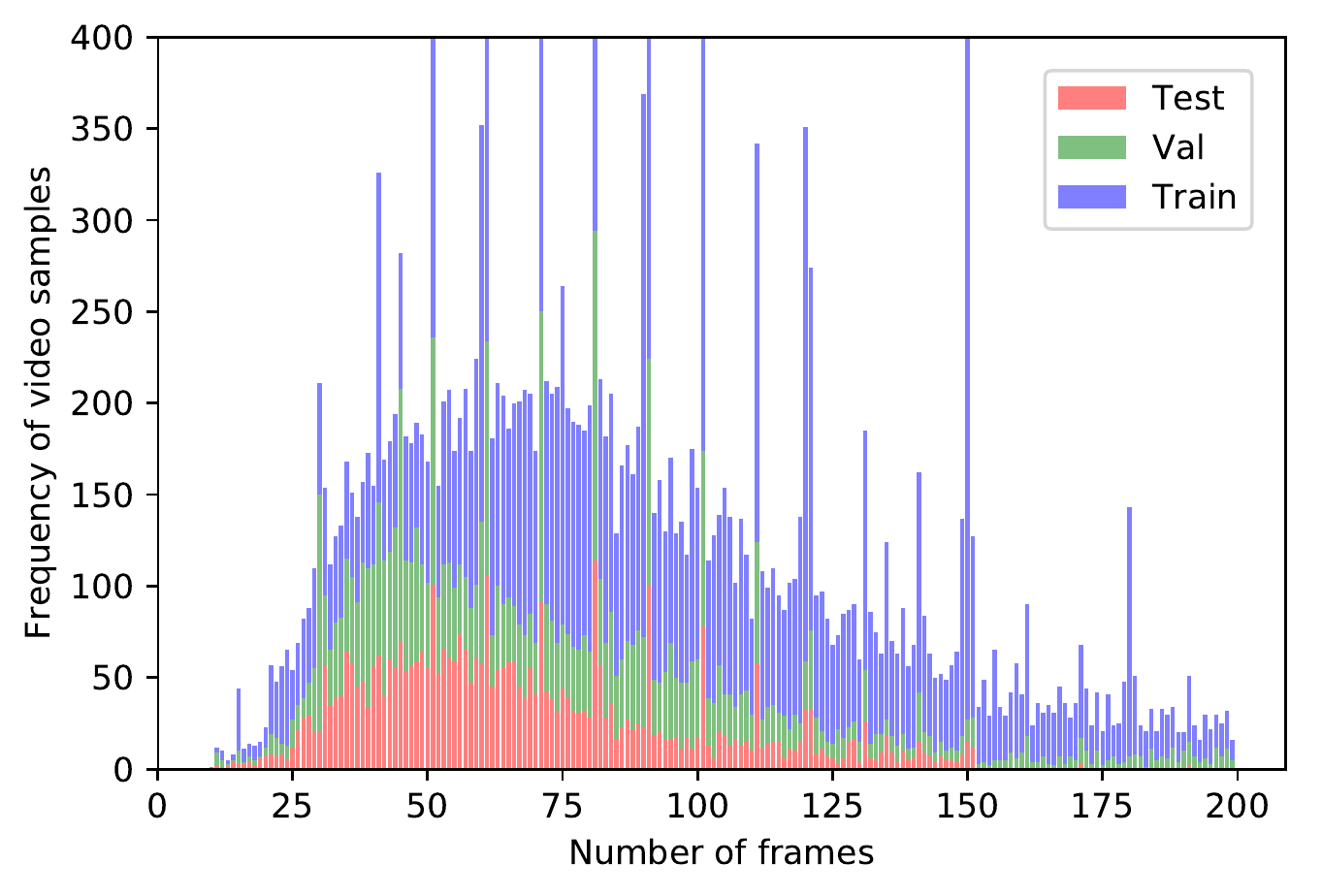}
   \vspace{-15pt}
   \caption{Histogram of frame numbers for ASL1000 video samples.} 
\label{fig:duration}
\end{minipage} \hfill
\begin{minipage}{0.5\textwidth}
   \centering
    \includegraphics[width=.85\linewidth]{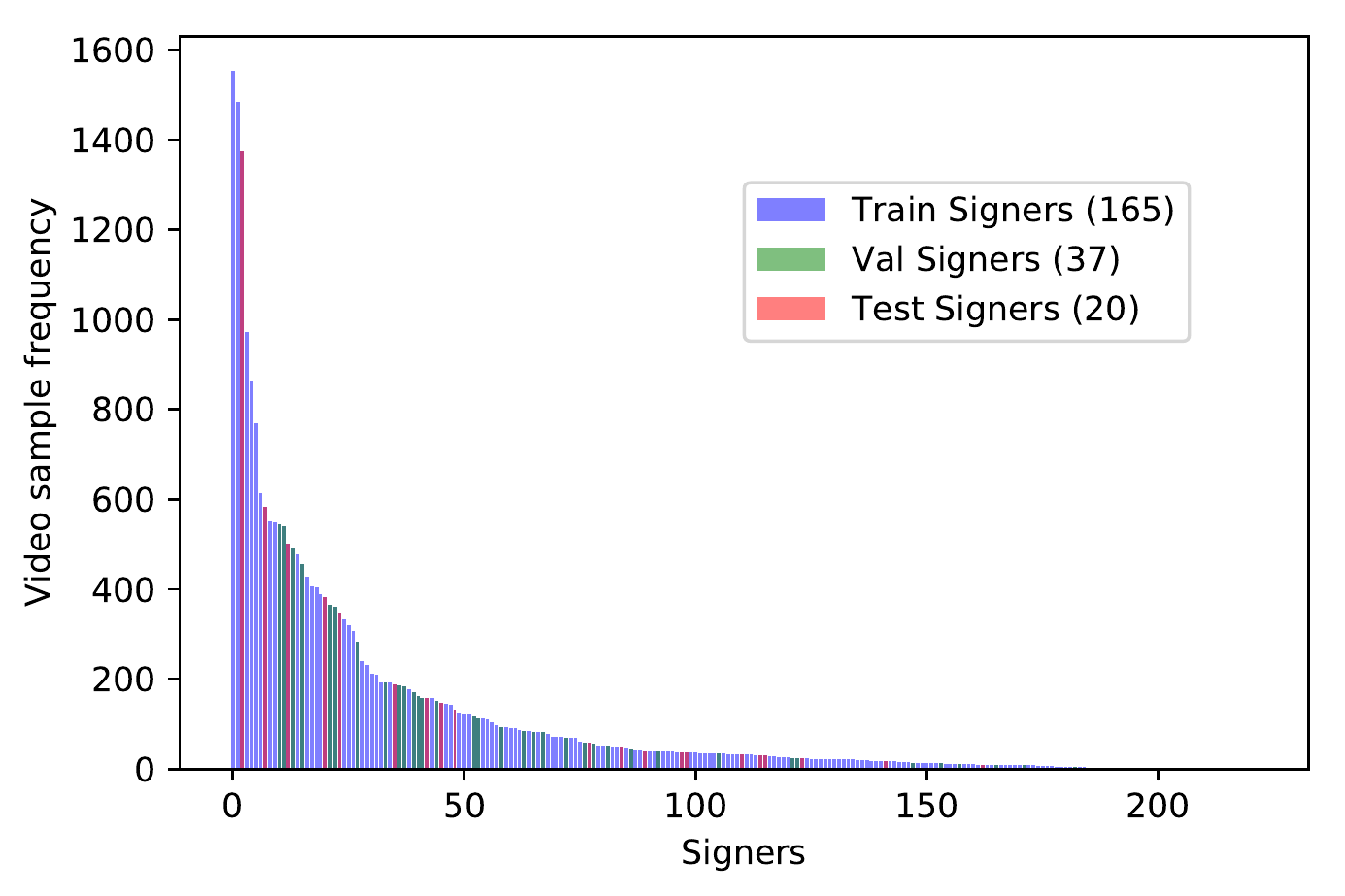}
    \vspace{-10pt}
   \caption{Showing the number of video samples for each of the 222 signers and the train/test/validation split of proposed data set.} 
\label{fig:signers}
\end{minipage}\hfill
\vspace{-10pt}
\end{figure}

Although many of the sample videos are good for training purposes,
some of them include the instruction to the sign or several repeated
performances with long pause in between. Therefore, we decided to
manually trim all video samples with a duration of more than 8
seconds. For higher accuracy on the test set, we chose the threshold
to be 6 seconds there. Although our annotators were not native in
ASL, they could easily trim these video samples while
considering other samples of the same label. We also decided to review
video samples shorter than 20 frames. In this way, around 25\% of the data set was manually
reviewed.
Figure~\ref{fig:duration} illustrates a histogram of the duration of
the 25,513 video samples of signs after the manual touch-up. There are
unusual peaks for multiples of 10 frames which seems to be caused by
video editing software cutting and adding captions, which favors such
duration. Despite that, the histogram looks like a Poisson
distribution with the average of 60. Combined, the duration of the
video samples is just over 24 hours long.  

\subsection{ASL synonyms}
Sign languages all over the world are independent, fully fledged
languages with their own grammar and word inventory, distinct from the
related spoken language. Sign languages can be characterized as
unwritten languages and have no commonly used written form. Therefore,
a written word will usually represent a semantic description and refer to the meaning of a sign, not to the way it is executed. This is
fundamentally different to most writing schemes of spoken languages. As an example, look at the two English
words \textit{Clean} and \textit{Nice}. While they are clearly
distinct in English, they have similar signs  in ASL which share
the same hand gesture and movement. On the other hand, the English
word \textit{Run} has a distinct sign for each of its meaning such as
"walk and run", "run for office", "run away" and "run a
business"~\cite{vicars2012american}. With respect to the ASL
videos we accessed from the internet and their descriptions, we needed
to make sure that similar ASL signs merged to one class for training
even if they have distinct English descriptors. This process was
implemented based on a reference ASL Tutorial
books~\cite{zinza2006master}. This mapping of sign classes will be
released as part of the MS-ASL data set.      

\subsection{Signer Identification}
Signer dependency is one of the most blocking challenges with current
non-intrusive sign recognition approaches. To address this issue, our
goal is to create a recognition corpus which covers signer independent
sets. We want to ensure that the signers occurring in train,
validation and test are distinct. Therefore, we aimed at identifying
the signer in each sample video. To achieve this, we computed 5 face
embeddings~\cite{schroff2015facenet} for each video sample. Based on
this, the video samples were then clustered into 457 clusters. Some of
these clusters were merged later by using the prior knowledge that two
consecutive samples from a video tend to have the same
signer. Additionally, we manually labeled the low confidence
clusters. Finally, we ended up having 222 distinct signers. The found
individuals occurred in the corpus with very diverse frequency. We
have 3 signers with more than one thousand video samples and 10
signers with a single video sample each. We then solve an optimization problem to distribute
signers into train, validation and test set signers aiming to divide
data set partitions to 80\%, 10\% and 10\% for train, validation and
test, respectively. However, due to the signer independency constraint
and unbalances samples, an exact division into these sizes was
impossible. We relaxed this condition, maintaining at least one sample
in each set for each class. The final amount of signers in each of the
sets was 165, 37 and 20 for train, validation and test,
respectively. Figure~\ref{fig:signers} shows the frequency of samples
by all 222 signers and the train/validation/test split.  

\subsection{MS-ASL Data Set with 4 Subsets}
In order to have a good understanding of the ASL vocabulary and being
a comprehensive benchmark for classifying signs with diverse training
samples,  
We release 4 subsets including 100, 200, 500 and 1000 most frequent
words. Each includes their own train, test and validation sets. All
these sets are signer independent and the signers for train (165),
test (20) and validation (37)  
are the same as shown in Figure~\ref{fig:signers}, therefore smaller sets are 
subset of the larger. We call these subsets \textit{ASL100},
\textit{ASL200}, \textit{ASL500} and \textit{ASL1000} for the rest of
this paper. 
Table~\ref{tab.sets} shows the characteristics of each of these
sets. In \textit{ASL100}, there are at least 45 samples for each class
while in \textit{ASL1000} there are at least 11 samples for each
class. 
\\

\textbf{Data Set Challenges:} There are challenges in this data
set which make it unique compared to other sign language data sets and more challenging compared to video
classification data sets: 
(1) One sample video may include repetitive act of a distinct signs. 
(2) One word can sign differently in different dialects based on geographical regions. As an example, there are 5 common signs for the word \textit{Computer}. 
(3) It includes large number of signers and is a signer independent data set.
(4)  They are large visual variabilities in the videos such as background, lighting, clothing and camera view point.

 \begin{table*}[t]
 \footnotesize
\begin{center}
\begin{tabular}{ l c c c c c c c c c}
\toprule
& & &  \multicolumn{4}{c}{\small{Number of Videos}} & Duration & \multicolumn{2}{c}{\small{Videos per class}}  \\
\cline{4-8} \cline{9-10} 
Data set & Class & Subjects & Train & Validation & Test & Total & [hours:min] & Min & Mean \\

\midrule

ASL100 & 100 & 189 & 3789 & 1190 & 757 & 5736 & 5:33 & 47 & 57.4 \\
ASL200 & 200 & 196 & 6319 & 2041 & 1359 & 9719 & 9:31 & 34 & 48.6  \\
ASL500 & 500 & 222 & 11401 & 3702  & 2720  & 17823 & 17:19 & 20  & 35.6  \\
ASL1000 & 1000 & 222 & 16054 & 5287 & 4172 & 25513 & 24:39 & 11 & 25.5  \\
\bottomrule
\end{tabular}
\vspace{-5pt}
\caption{Showing statistics of the 4 proposed subsets of the MS-ASL data set.} \label{tab.sets}
\end{center}
\vspace{-10pt}
\end{table*}
\subsection{Evaluation Scheme}
\label{sec:evaluationscheme}

We suggest two metrics for evaluating the algorithms ran in these data
sets: 1) average per class accuracy, 2) average per class top-five
accuracy. We prefer per class accuracy compared to plain accuracy to
better reflect performance given the unbalance test set inherited from the unbalance nature of the data
set. To be more precise, we compute the
accuracy of each class and reported the average value. In the top-5
accuracy, we call it correct if the ground-truth label appears in the
top five guesses of the method being evaluated. We compute top-five
accuracy for each class and report the average value. ASL, just like any other language
can have ambiguity which can be resolved in context. Therefore, we picked top-five accuracy. 
%
\section{Baseline Methods}
\label{sec:method}
Although it is much more challenging, but we can consider isolated
sign language recognition similar to action recognition or gesture
detection as it is a video classification task for a human being. We
can categorize current action recognition or gesture detection into
three major categories or combination of them 1) Using 2D convolution
on image and do a recurrent network on top of that
~\cite{donahue2015long, feichtenhofer2016convolutional} 2) Extracting
subject's body joints in the form of skeleton and using skeleton data
for recognition~\cite{du2015hierarchical, zhu2016co} 3) Using 3D
convolution~\cite{carreira2017quo, tran2015learning,
  molchanov2016online}. In order to have baselines from each
categories of human action recognition, we implement at least one
method for each of these categories.  

For all of the methods, we use the bounding box covering the signer as input
image. We extract the person bounding box by the SSD
network~\cite{liu2016ssd} and release it for each video sample as part
of MS-ASL data set. 
For spatial
augmentations, body bounding boxes are randomly scaled or translated
by 10\%, fit into a square and re-sized to fixed $224\times224$
pixels. We picked $64$ as our temporal window which is the average number
of frames across all sample videos. In addition, the resulted
video is randomly but consistently (per video) flipped horizontally because ASL is
symmetrical and can be performed by either hands.  We used fixed sized
frame number as well as fixed size resolution for 2D and 3D
convolution methods. For temporal augmentations: $64$ consecutive
frames are picked randomly from the videos and shorter videos are
randomly elongated by repeating their fist or last frame. We train for 40
epochs. 
In this paper, we focused on RGB only algorithms and did not use
optical flow for any of the implementations. It is a proven fact that
using optical flow as second stream in train and test
stage~\cite{simonyan2014two,feichtenhofer2016convolutional,cui_deep_2019} or just
train stage~\cite{abavisani2018unimodal} boosts the performance of
prediction. Herein, we describe the methods used for determining
baselines.  
\\

\textbf{2D-CNN : }
The high performance of 2D convolutional networks on image
classification makes them the first candidate for video
processing. This is achieved by extracting features from each frame of
the video independently. The first approach was to combine these features
by simply pooling the predication, but it ignored the frame ordering
or timing. The next approach which proved more successful, was using
recurrent layers on the top of 2D convolution networks. Motivated by
\cite{donahue2015long}, we picked LSTM~\cite{hochreiter1997long} as
our recurrent layer which records the temporal ordering and long range
dependencies by encoding the states. We used
VGG16~\cite{simonyan2014very} network followed by an average pooling
and LSTM layer of size 256 with batch normalization. The final layers
are a 512 hidden units followed by a fully connected layer for
classification. We considered the output on final frame for testing.
We also have implemented \cite{koller2017:re-sign} as the state-of-the-art
on PHOENIX2014 data
set~\cite{forster_extensions_2014,koller_continuous_2015}. This method
use GoogleNets~\cite{43022} as 2D-CNN with 2  bi-directional LSTM
layers and 3-state HMM. We report it as Re-Sign in the experimental
result.  
\\

\begin{figure}
\begin{center}
   \includegraphics[width=0.11\linewidth]{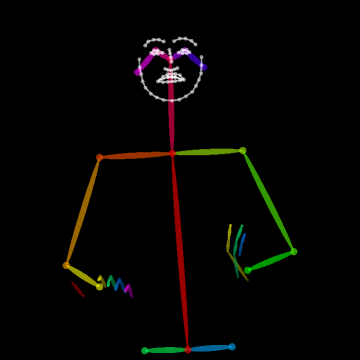}
   \includegraphics[width=0.11\linewidth]{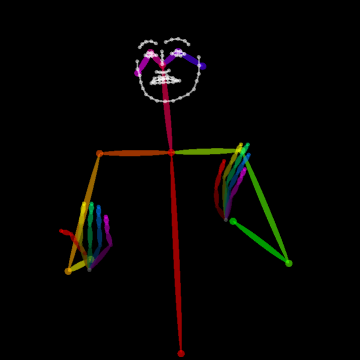}
   \includegraphics[width=0.11\linewidth]{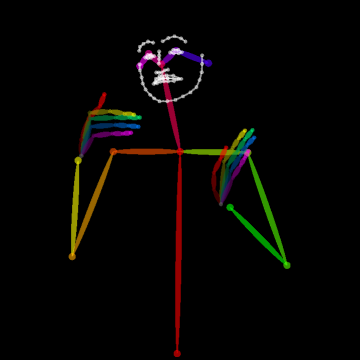}
   \includegraphics[width=0.11\linewidth]{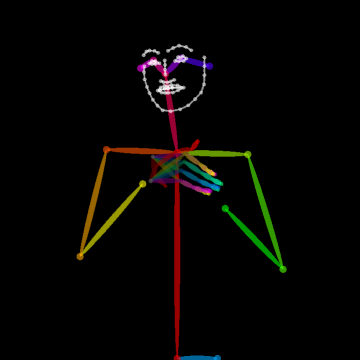}
   \includegraphics[width=0.11\linewidth]{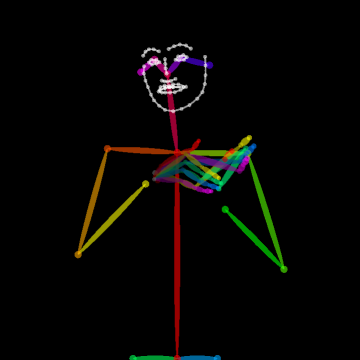}
   \includegraphics[width=0.11\linewidth]{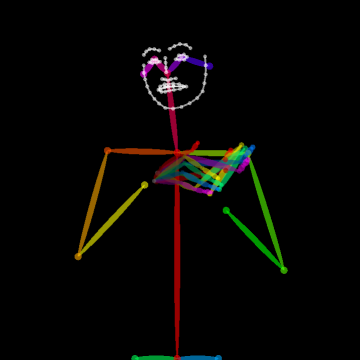}
   \includegraphics[width=0.11\linewidth]{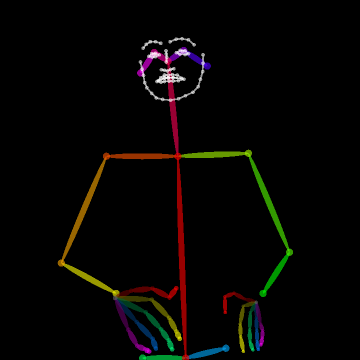}
   \includegraphics[width=0.11\linewidth]{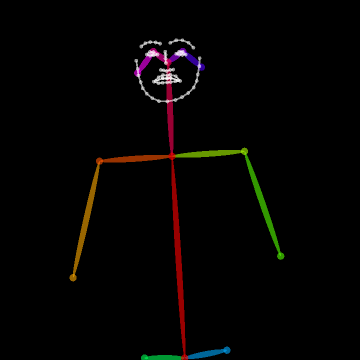}
\end{center}
\vspace{-15pt}
   \caption{Extracted 137 body key-points for a video sample from MS-ASL by~\cite{cao2017realtime, simon2017hand}. }
\label{fig:sk}
\vspace{-10pt}
\end{figure}

\textbf{Body Key-Points : }
With the introduction of robust body key-points (so-called skeleton)
detection~\cite{wei2016cpm}, some studies try to solve human action
recognition by body joints only~\cite{du2015hierarchical, zhu2016co}
or use body joints along with the image
stream~\cite{choutas2018potion}. Since most body key-point techniques
did not cover hand details, it was not rational to use it for sign
language recognition task as it relies heavily on the movement of
fingers. But a recent work has covered hand and face key-points along
with classical skeleton~\cite{simon2017hand}. We leveraged this
technique which extracted 137 key-points in total, to do a baseline on
our data set by body key-points. We extracted all the key-points for
all samples using ~\cite{cao2017realtime, simon2017hand}.  Using 64
frames for time window, our input to the network would be
$64\times137\times3$ representing $x$, $y$ coordinates and confidence
values for the 137 body key-points for all consecutive 64
frames. Figure~\ref{fig:sk} illustrates the extracted 137 body
key-points for a video sample from MS-ASL. 
The hand key-points are not as robust as body and face. 

 We implemented hierarchical co-occurrence network (HCN)~\cite{zhu2016co} which originally used 15 joints. We extended this work by using 137 body key-points including hand and face key-points. The input to this network is original 137 body key-points as well as per frame difference of them. The network includes three layers of 2D convolution on top of each input as well as two extra 2D convolution layers after the concatenation of two paths. 
\\

\textbf{3D-CNN : }
Recently, 3D convolutional networks have shown promising performance
for video classification and action recognition including C3D
network~\cite{tran2015learning} and I3D
network~\cite{carreira2017quo}. 
We applied C3D~\cite{tran2015learning} released code from author as
well as our own implemented version to our proposed data sets with and
without pre-trained model, trained on
Sport-1M~\cite{karpathy2014large}. The model did not converge for any
of our experiments.     
We adopted the architecture of I3D networks proposed
in~\cite{carreira2017quo} and employed its suggested implementation
details.  This network is an inflated version of
Inception-V1~\cite{ioffe2015batch}, which contains several 3D
convolutional layers followed by 3D max-pooling layers and inflated
Inception-V1 submodules. We started with pre-trained network trained
on Imagenet~\cite{deng2009imagenet} and
Kinetics~\cite{carreira2017quo}. We optimized the objective functions
with standard SGD with momentum set to 0.9. We began the base learning
rate at $10^{-2}$ with a $10\times$ reduction at epoch 20 when
validation loss saturated.   

\section{Experimental Result}
\label{sec:result}

We trained all of the methods mentioned in section~\ref{sec:method} on
four MS-ASL subsets (\textit{ASL100, ASL200, ASL500} and
\textit{ASL1000}) and computed the accuracy for test set which
includes subjects that are not included in training phase. As
described in subsection~\ref{sec:evaluationscheme}, we report two
evaluation metrics: average per class accuracy and average per class
top-five persent accuracy. The results are reported in Table~\ref{tab:result}
and Table~\ref{tab:result-top5} respectively. We did not over optimize
training parameters.
Hence, these results constitute baseline for
2D-CNN, 3D-CNN, CNN-LSTM-HMM and body key-point based approaches. The experimental
result suggests that this data set is very difficult for 2D-CNN or at
least LSTM could not propagate the recurrent information well. In video
classification data sets such as UCF101~\cite{soomro2012ucf101} or
HMDB51~\cite{kuehne2013hmdb51}, the image itself carries context
information regarding the classification while in MS-ASL there is
minimum context information in a single image. Re-Sign~\cite{koller2017:re-sign} which report as state-of-the-art in few sign language 
dataset could not achieve well for challenging MS-ASL (This method could not predict top-five). Body key-point based
approach (HCN) is doing relatively better compared to 2D-CNN but there
is huge room for improvement because of network simplicity as well
as future improvements for hand key-point extraction. On the other
hand our 3D-CNN baseline did achieve good results in this
challenging, uncontrolled data set and we propose it as powerful
network for sign language recognition.  

\begin{table}[tb]
\small
\begin{center}
   \renewcommand{\arraystretch}{0.9}
  \setlength{\tabcolsep}{4pt}
\begin{tabular}{ l c c c c}
\toprule
Method                      & \small{ASL100} & \small{ASL200} & \small{ASL500} & \small{ASL1000} \\
\midrule
Naive Classifier    & 0.99 & 0.50 & 0.21 & 0.11 \\  
VGG+LSTM \cite{cui_recurrent_2017,cui_deep_2019}                & 13.33        & 7.56         & 1.47         & -               \\  
HCN \cite{zhu2016co} & 46.08        & 35.85        & 21.45        & 15.49         \\
Re-Sign \cite{koller2017:re-sign} & 45.45        & 43.22          &27.94      & 14.69         \\
I3D \cite{carreira2017quo} & \textbf{81.76}        & \textbf{81.97}        & \textbf{72.50}        & \textbf{57.69}         \\
  \bottomrule
\end{tabular}
\vspace{3pt}
\caption{The average per class accuracy for baseline method on proposed ASL data sets.} 
\label{tab:result}
\end{center}
\vspace{-14pt}
\end{table}

\begin{table}[h]
\small
\begin{center}
\begin{tabular}{ l c c c c}
\toprule
Method & \small{ASL100} & \small{ASL200} & \small{ASL500} & \small{ASL1000} \\
  \midrule
  Naive classifier & 4.86  & 2.49  & 1.05  & 0.58  \\
VGG+LSTM \cite{cui_recurrent_2017,cui_deep_2019}& 33.42  & 21.21  & 5.86  & - \\
HCN~\cite{zhu2016co}  & 73.98 & 60.29 & 43.83 & 32.50 \\
I3D~\cite{carreira2017quo}  & \textbf{95.16} & \textbf{93.79} & \textbf{89.80} & \textbf{81.08} \\
  \bottomrule
\end{tabular}
\vspace{3pt}
\caption{ The average per class top-five accuracy for baseline methods on proposed data sets. } 
\label{tab:result-top5}
\end{center}
\vspace{-14pt}
\vspace{-10pt}
\end{table}

\subsection{Qualitative Discussion}
\label{res.one}

\begin{figure}[t]
\begin{minipage}{0.45\textwidth}
   \centering
    \includegraphics[width=.95\linewidth]{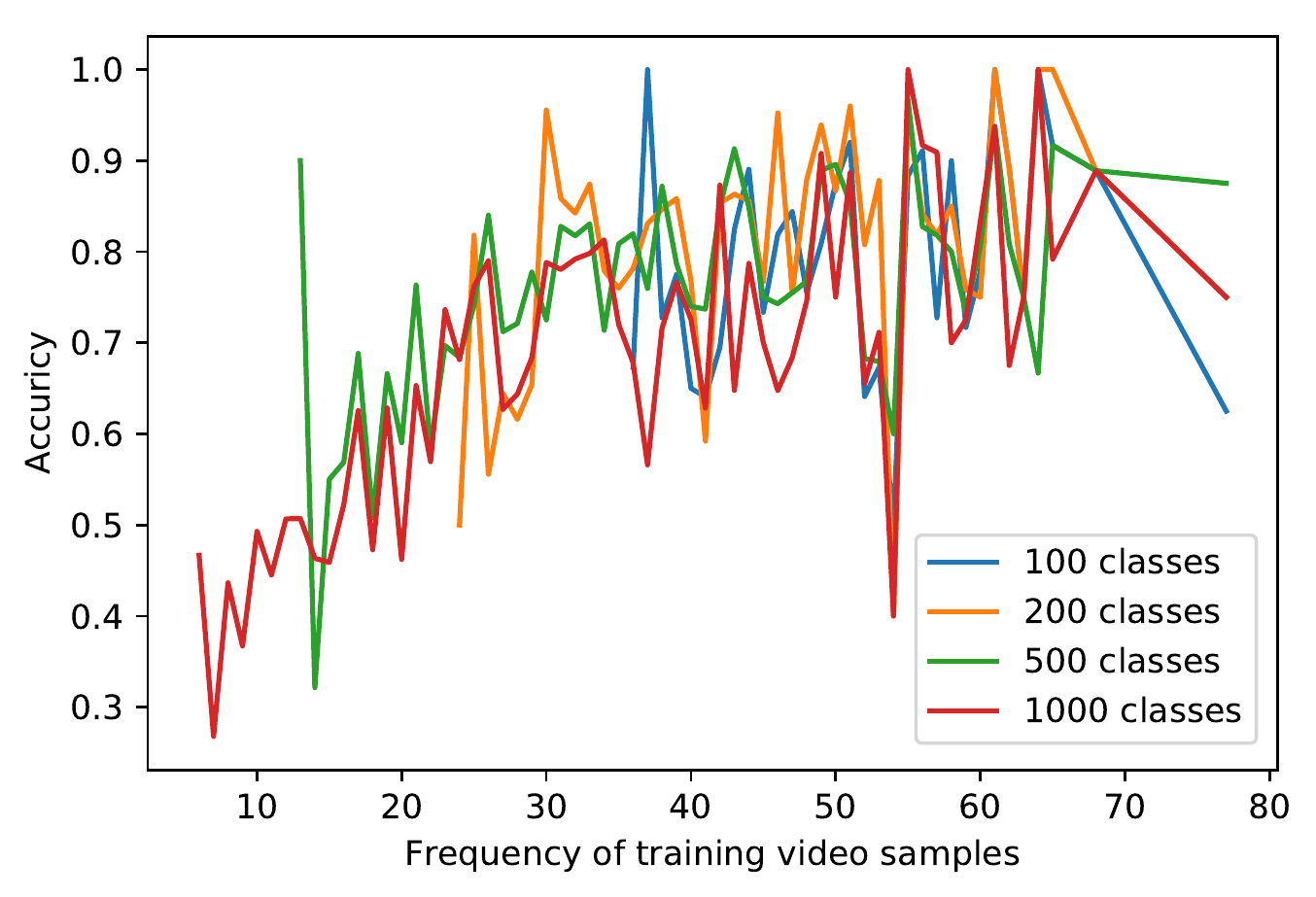}
    \vspace{-10pt}
   \caption{The accuracy of trained models based on frequency of training samples.} 
\label{fig:acc}
\end{minipage}\hfill
\begin{minipage}{0.45\textwidth}
   \centering
   \includegraphics[width=0.8\linewidth]{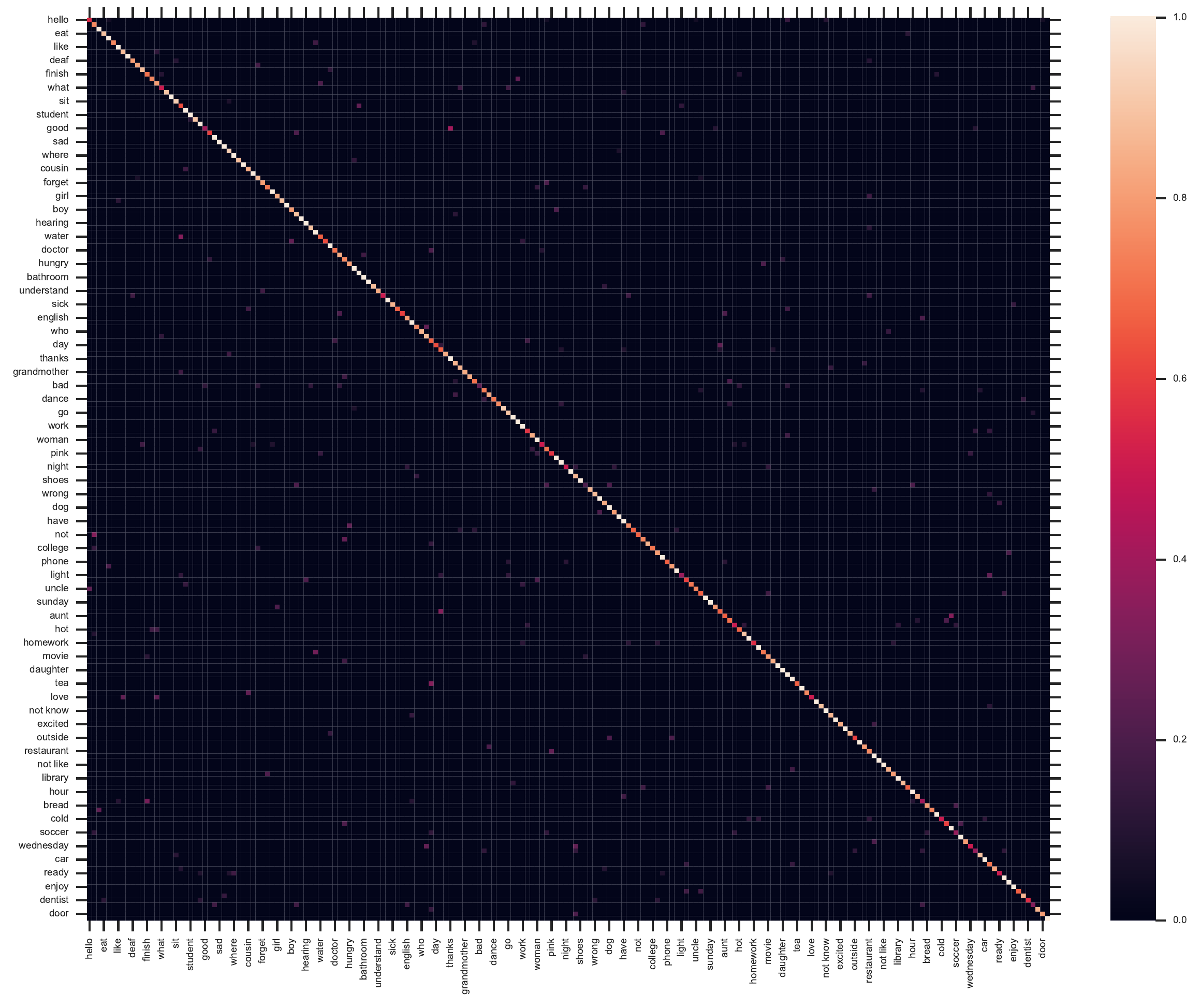}
   \vspace{-9pt}
   \caption{The confusion matrix for I3D on \textit{ALS200} data set.} 
\label{fig:confusionmatrices}
\end{minipage} \hfill
\vspace{-10pt}
\end{figure}

Figure~\ref{fig:confusionmatrices} illustrates the confusion matrix
obtained by comparison of the grand-truth labels and the predicted
labels from models trained by I3D on \textit{ALS200} data set. As we
expected, most of the values lay on the diagonal element. Here is the
list of brightest points off the diagonal with value of more than .25
which represents per class worst predictions: 

- \textit{Good} labeled as \textit{Thanks} (.4): often the sign \textit{Good} is done without the base hand, this sign can mean \textit{Thanks} or \textit{Good} 

- \textit{Water} labeled as \textit{Mother} (.33): both by placing dominant hand around chin area while the detail is different. 

- \textit{Today} labeled as \textit{Now} (.33): two versions for \textit{Today} one of them is signing \textit{Now} twice.  

- \textit{Not} labeled as \textit{Nice} (.33), \textit{Aunt} labeled as \textit{Nephew} (.33), \textit{Tea} labeled as \textit{Nurse} (.33)

- \textit{Start} labeled as \textit{Finish} (.3)

- \textit{My} labeled as \textit{Please} (.28): both sign by place the dominant hand on the chest. A clockwise motion for \textit{Please} and gentle slapping for \textit{My}.  

We did similar investigation for other data sets and find interesting evidence about language ambiguity that could solve within the context. 
Therefore, the error of the model is combination of language ambiguity
and prediction error. Our observation shows when we have smaller
training sets, model error mainly come from prediction errors but for
classes with more samples the error could came from language
ambiguity. 
This advise us to use five-top as our second metric since eventually
these predication need to feed to language model with context.

\subsection{The Effect of Pre-Trained Model}
The fact that I3D training on \textit{ASL200} outperformed I3D trained
on \textit{ASL100} was not convincing as it contains twice the classes
as \textit{ASL100}. We verified this result with further
experiments. We evaluate the I3D model trained with the
\textit{ASL200} data on \textit{ASL100} to study the effect of data.
The average per class accuracy reaches
$83.36\%$ which made the results less convincing. The only proposed
explanation is the lack of adequate training video samples which is
less than four thousands. This prompted us to do a new experiment; We
trained I3D on \textit{ASL100} using the same setting as the last
experiments except for using \textit{ASL200} as pre-trained model
instead of ImageNet+Kinetics pre-trained model. The result was
$85.32\%$ for average per class accuracy and $96.53\%$ for average per
class top-five accuracy which is more than $3.5\%$ performance boost.
This is a valid experimental approach as the test and train are still
separated due to signer independency. This verifies our reasoning and
suggests that the existing out-of-domain pre-trained models can be easily
outperformed by in-domain pre-trained models specific for sign
language recognition. We proposed the model trained on MS-ASL as
a I3D pre-trained model for sign language recognition tasks.           

\subsection{ The Effect of Number of Classes}
In order to evaluate the effect of number of classes in model
prediction, we tested the I3D model trained on \textit{ASL1000}
training sets on \textit{ASL500, ASL200} and \textit{ASL100} test
sets. This allowed a comparison between the model trained on 100
classes with the one trained with 1000 classes on the same test
set. We did similar experiments with all possible pairs and reported
the average per class accuracy on Table~\ref{tbl:increase}. In this
table we show subsets of the MS-ASL data set on the horizontal axis
and the tested subsets on the vertical axis. Increasing the number of
classes decreased the accuracy of either the train or the test
phase. Doubling the size of test classes led to a small change from
$83.36\%$ to $81.97\%$ and doubling the size of the train classes from
$85.32\%$ to $83.36\%$. This suggests that the observed effect is
significantly less when we have more video samples per class.

\begin{table}[h]
\small
\begin{center}
\begin{tabular}{ l c c c c}
\hlineB{2}
\hline
\small{I3D trained on} & \small{ASL100} & \small{ASL200} & \small{ASL500} & \small{ASL1000} \\
\hline\hline
ASL100   & 85.32\%  & - & - & - \\
ASL200   & 83.36\% & 81.97\% & - & - \\
ASL500  & 80.61\% & 78.73\% & 72.50\% & - \\
ASL1000  & 75.38\% & 74.78\% & 68.49\% & 57.69\% \\
\hlineB{2}
\end{tabular}
\vspace{2pt}
\caption{ Showing the average per class accuracy of the model trained on different subsets of the MS-ASL data set (horizontal), subsets tested on (vertical).} 
\label{tbl:increase}
\end{center}
\vspace{-18pt}
\end{table}

\subsection{The Effect of Number of Video Samples}

In order to determine the adequate number of video samples per word
needed to a good model, we experimented with the number of samples
illustrated figure~\ref{fig:acc}. It shows the accuracy of the models
based on frequency of training data for our experiments on test
data. It shows a somewhat similar curve for all the four experiments
suggesting that the accuracy correlates directly to the number of
training video samples for classes with less than 40 video
samples. However, for classes with more than 40 video samples, the
difficulty of the signs may be more important. Although we have
average accuracy of $80\%$ for classes with more than 40 training
video samples, it does not suggest that 40 is the sweet spot. Direct
comparison cannot be made as this dataset lacks other classes which
are significantly larger than 40 video samples. The curve deep at
$x=54$ for all networks belongs to the class \textit{Good} which is
the only class with 54 training samples. We have discussed this in
subsection~\ref{res.one}.   

\vspace{-10pt}
\section{Conclusion}
\label{sec:conclusion}
In this paper, we proposed the first large-scale ASL data set
with $222$ signers and signer independent sets. Our dataset contains a
large class count of $1000$ signs recorded in challenging and
unconstrained conditions. We evaluated the state-of-the-art network
architectures and approaches as the baselines on our data set and
demonstrated that I3D outperforms current state-of-the-art methods by
a large margin. We also
estimated the effect of number of classes on the recognition accuracy.   

For future works, we propose applying optical flow on the videos as it
is a strong information extraction tool. We can also try leveraging
body key-points and segmentation on the training phase only. We
believe that the introduction of this large-scale data set will
encourage and enable the sign language recognition community to catch
up with latest computer vision trends.

{\small
\bibliography{egbib,thesis-koller}
}

\end{document}